\documentclass{ifacconf}

\usepackage{graphicx}      %
\usepackage{natbib}        %
\usepackage{amsmath}
\usepackage{mathtools} %
\usepackage{amssymb} %
\usepackage{color}
\usepackage{algorithm}
\usepackage{algpseudocode}
\usepackage{caption}
\usepackage{longtable} 
\usepackage{booktabs}

\usepackage{url}
\usepackage[T1]{fontenc}
\usepackage{upquote}
\usepackage{commath}
\usepackage{setspace}
\usepackage[table,xcdraw]{xcolor}
\usepackage{multirow}
\usepackage{makecell}
\usepackage{stackengine}

\AtBeginEnvironment{thebibliography}{\linespread{0.95}\selectfont}

\usepackage{flushend}

\RequirePackage[absolute]{textpos}
\DeclareRobustCommand*{\copyrightnote}[1]{%
  \begin{textblock}{200}(15,286.2)
      \footnotesize #1%
  \end{textblock}%
}

\usepackage{soul} %
\input{abkuerzung.sty}
\setstcolor{green}

\DeclareMathSizes{10}{9}{7}{6}

\begin{document}
\begin{frontmatter}

\title{%
Fast yet predictable braking manoeuvers for real-time robot control$^\star$
} 

\thanks[footnoteinfo]{%
The research leading to these results has received funding from the European Union's Horizon 2020 research and innovation program as part of the project DARKO (grant no. 101017274) and the Marie Skłodowska-Curie action (grant no. 899987). The authors would like to thank the Bavarian State Ministry for Economic Affairs, Regional Development, and Energy (StMWi) for financial support as part of the project SafeRoBAY (grant no. DIK0203/01). This work was also supported by the German Research Foundation (DFG) as part of Germany’s Excellence Strategy, EXC 2050/1, Project ID 390696704 – Cluster of Excellence “Centre for Tactile Internet with Human-in-the-Loop” (CeTI) of Technische Universität Dresden.  
}

\author[First]{Mazin Hamad} 
\author[First]{Jesus Gutierrez-Moreno}
\author[First]{Hugo T. M. Kussaba}
\author[Second]{Nico Mansfeld} 
\author[First]{Saeed Abdolshah}
\author[First,Third,Fifth]{Abdalla Swikir}
\author[Fourth]{Wolfram Burgard}
\author[First,Fifth]{Sami Haddadin}

\address[First]{Chair of Robotics and Systems Intelligence, Munich Institute of Robotics and Machine Intelligence, Technical University of Munich, Germany. (e-mail: mazin.hamad@tum.de).}
\address[Second]{Franka Emika GmbH, Munich, Germany.}
\address[Third]{Department of Electrical and Electronic Engineering, Omar Al-Mukhtar University, Libya.}
\address[Fourth]{Department of Engineering, University of Technology Nuremberg, Germany.}
\address[Fifth]{Centre for Tactile Internet with Human-in-the-Loop (CeTI), Germany.}

\begin{abstract}                %
This paper proposes a framework for generating fast, smooth and predictable braking manoeuvers for a controlled robot. The proposed framework integrates two approaches to obtain feasible modal limits for designing braking trajectories. The first approach is real-time capable but conservative considering the usage of the available feasible actuator control region, resulting in longer braking times. In contrast, the second approach maximizes the used braking control inputs at the cost of requiring more time to evaluate larger, feasible modal limits via optimization. Both approaches allow for predicting the robot’s stopping trajectory online. In addition, we also formulated and solved a constrained, nonlinear final-time minimization problem to find optimal torque inputs. The optimal solutions were used as a benchmark to evaluate the performance of the proposed predictable braking framework. A comparative study was compiled in simulation versus a classical optimal controller on a 7-DoF robot arm with only three moving joints. The results verified the effectiveness of our proposed framework and its integrated approaches in achieving fast robot braking manoeuvers with accurate online predictions of the stopping trajectories and distances under various braking settings. 

\end{abstract}


\end{frontmatter}

\copyrightnote{\copyright 2023 the authors. This work has been accepted to IFAC for publication under a Creative Commons Licence CC-BY-NC-ND.}

\section{Introduction}
Collaborative robots, also known as cobots, have become increasingly popular in manufacturing applications in recent years \citep{matheson2019human}. These robots enable human-robot collaboration (HRC) in shared workspaces without the need for cages or fences to separate human and robot. 
This is especially important in recent high mix/low volume production scenarios, where a fenceless operation allows highly reconfigurable and adaptable cell designs enabling several production flows to be handled concurrently \citep{schlette2020towards}.
Before deploying a cobot into a fenceless, shared workspace with dynamically moving objects, it must be ensured that they are always fully capable of reacting to the motions of nearby obstacles and human coworkers in a compliant and safe manner. 

For applications involving direct HRC or physical human-robot interaction (pHRI), safety standards are decisive for both industrial and service robots. Such safety standards specify strict requirements and guide the robot's mechanical design, task planning, and motion control to ensure safe pHRI. More specifically, technical specifications, such as e.g. the TS 15066,\footnote{Robots and robotic devices — Collaborative robots (ISO/TS 15066:2016) \cite{TS15066}.} supplement these standards and define four safeguarding modes for collaborative operation. For instance, any contact with the human head is strictly forbidden, and the TS15066 requires implementing a safety-rated, monitored stop (SMS) to stop the robot through control. This means that the integrated SMS mode must always stop the robot upon detecting faulty events (\eg colliding with obstacles) or predicting non-safe, poorly coordinated movements of the human coworker that may result in any injury risk. For this, the ISO 13850 standard\footnote{Safety of machinery — Emergency stop function — Principles for design (ISO 13850:2015) \cite{ISO13850}.} defines three types of stop functions

\begin{itemize}
    \item \textbf{Category 0 stop} corresponds to the immediate removal of power from the robotic system actuators (i.e., uncontrolled stop).
    \item \textbf{Category 1 stop} is meant to be a controlled stop, but still, the removal of power occurs when the stop is achieved.
    \item \textbf{Category 2 stop} corresponds to a controlled stop with power kept available to the system actuators. 
\end{itemize}
This means both Category 0 and Category 1 stops inevitably require power removal upon emergencies, which may later be followed by human intervention to restart the system and its operation. Hence, for a more efficient collaborative task, Category 2 stop is needed. However, without a good braking strategy, executing Category 2 stops during the robot control cycle is impossible. 

\begin{figure}[t]
    \centering
    \includegraphics[width=0.8\columnwidth]{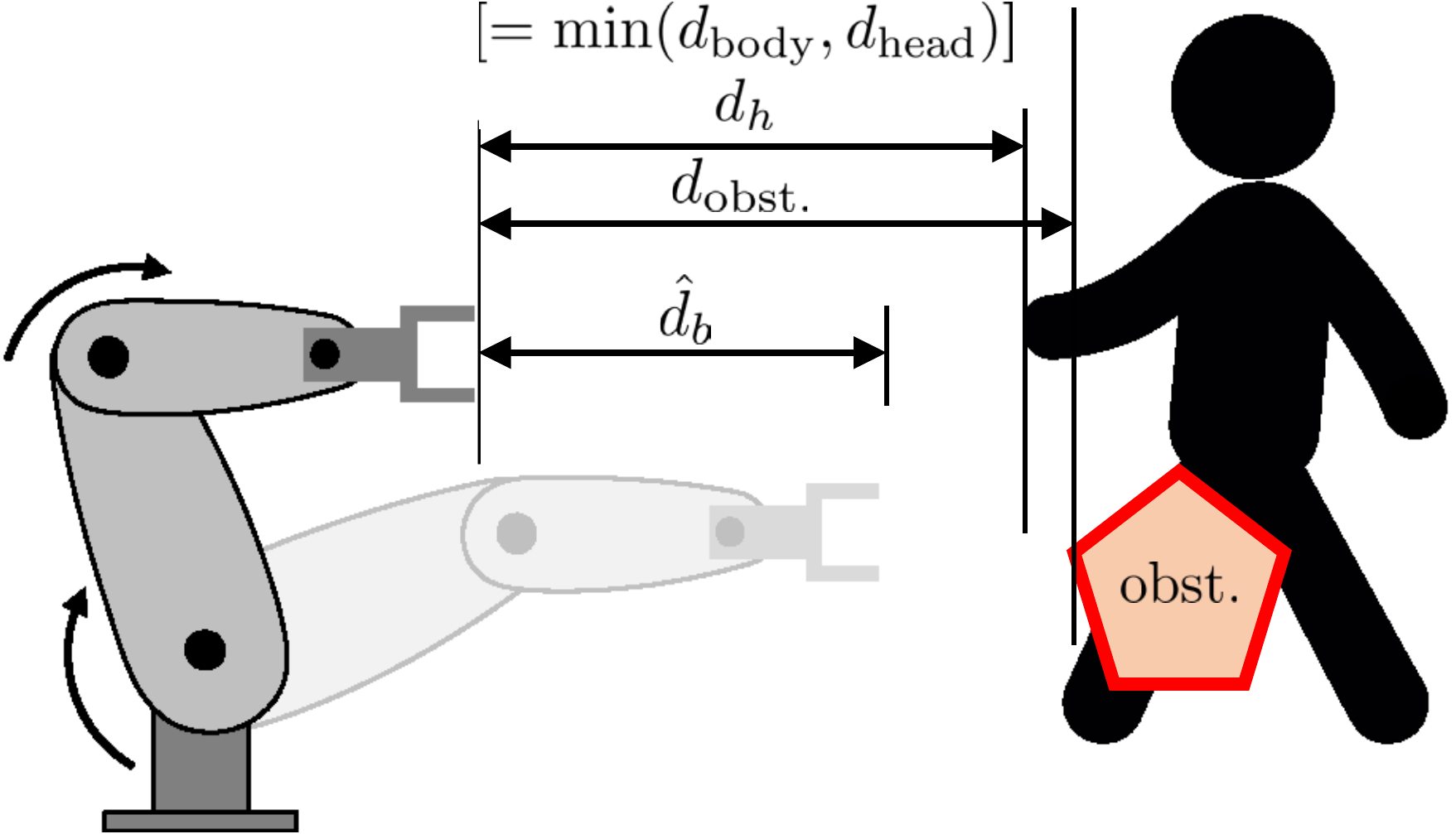}
    \caption{A conceptual pHRI scenario. A robot arm executes its task in a fenceless, shared workspace with human coworkers. Since undesired collisions may occur, the robot must be equipped with controlled stops to be triggered when its current braking distance $\hat{d}_b$ drops below dynamically evaluated distance thresholds. The scalar quantities $\hat{d}_h,\,\hat{d}_{obst.}$ denote the shortest distances to the closest human body parts and obstacles in the robot vicinity, respectively.} %
    \label{fig:title_figure}
\end{figure}

Ideally, the robot performs its tasks at high speed to deliver a high production throughput when working autonomously. To adhere to safety constraints in shared workspaces or collaborative regimes, the robot may have to reduce its performance or even stop immediately by switching to some safe functional mode \citep{svarny2022functional} in case of any potential collision with a human. This requires estimating the distance traveled by the end-effector $EE$ from the braking instant until the robot is brought to a complete stop (the so-called \textit{robot stopping distance}), see Fig.~\ref{fig:title_figure}. Applications involving close pHRI (i.e., active collaboration) require a smooth robot braking with predictable distances, so it is possible to evaluate relative distances between the robot and dynamic obstacles (including humans) in its vicinity. Achieving this provides a rigorous braking solution emphasizing human safety in direct pHRI. 
In this manuscript, we propose a braking framework capable of stopping the motion of a robotic system smoothly within its actuation capabilities. This is achieved by designing 
smooth braking trajectories for the velocities after transforming the system dynamics into a modal space representation whose decoupled coordinates can be controlled independently. The designed modal braking trajectories are transformed back to the original space to generate the required robot braking manoeuvers, which are applied as control inputs at each time step until the complete stop. The proposed braking solution does not require using any external hardware or additional sensors other than the robot's joint actuators and their encoders.\footnote{Of course, a braking trigger has to be integrated, which is inevitably connected to additional tracking hardware such as, e.g., a lidar and/or an RGB-D camera for monitoring the collaborative workcell.} The presented framework integrates approaches for generating fast braking manoeuvers in real-time (\ie it finishes all the required computations, including braking control and stopping trajectory prediction algorithms, in one control cycle). It further generalizes to robotic systems with arbitrary degrees of freedom (DoFs), making it scalable to various types of complex robotic systems.

This manuscript is structured as follows. Section~\ref{sec:probelm_modelling} presents the modeling of the considered problem. The proposed braking scheme is detailed in Sec.~\ref{sec:proposed_scheme}. Simulation results are provided in Sec.~\ref{sec:results}, including comparisons against a classic solution obtained by optimal control approaches and their evaluations. Section~\ref{sec:conclusion} concludes the paper and highlights future research directions. 

\section{Modelling of the problem}
\label{sec:probelm_modelling}

\subsection{Robot model}
The considered robotic system is an $n$-DoF rigid robot arm that has only
revolute joints and is modeled by the non-linear differential equations
\begin{equation}
    \vM(\vq) \ddot{\vq} + \vC(\vq, \dot{\vq}) \dot{\vq} + \vtau_{f} + \vg(\vq) = \vtau_J + \vtau_\mathrm{ext},
    \label{eq:robot_dynamics}
\end{equation}
where $\vq, \dot{\vq} \in \mathbb{R}^{n}$ ($n$ is the number of joints) are, respectively, the joint position (generalized coordinates) and velocity vectors, which constitute together the state of the arm at a given instant in time.\footnote{To avoid a cluttered notation, we have left out the dependence of the variables on time $t$.} The joint acceleration vector is denoted $\ddot{\vq} \in \mathbb{R}^{n}$, $\vM(\vq) \in \mathbb{R}^{n \times n}$ is the inertia matrix, $\vg(\vq) \in \mathbb{R}^n$ is the gravity torque vector and $\vC(\vq, \dot{\vq}) \dot{\vq}=$ $\vc(\vq, \dot{\vq}) \in \mathbb{R}^n$ is the vector of the Coriolis and centrifugal forces, with the matrix $\vC(\vq, \dot{\vq})$ defined through the Christoffel symbols of the first kind, satisfying $\dot{\vM}(\vq)=\vC(\vq, \dot{\vq})+\vC^{\tp}(\vq, \dot{\vq})$ \citep{yin1989efficient}. %
The vector of joint friction torques is denoted $\vtau_{f} \in \mathbb{R}^n$. The torques $\vtau_J \in \mathbb{R}^n$ produced by the joint motors are input to the system, while $\vtau_\mathrm{ext} = \vJ^{\tp}(\vq) \vf_\mathrm{ext}$ are the external torques exerted on the arm's end-effector by the environment, where $\vJ(\vq) \in \mathbb{R}^{p \times n_i}$ is the arm Jacobian matrix and $\vf_\mathrm{ext} \in \mathbb{R}^p$ is the vector of external forces.

\begin{assum}\label{assum:compensated_model}
The gravity and Coriolis torques are considered quasi-stationary over the braking trajectory since the braking is a local behavior of the robot system \citep{mansfeld2014reaching}. Also, the non-conservative forces, mainly from the gearing friction, are negligible because of the high-bandwidth low-level controller based on feedback from the joint actuators \citep{terry2017comparison}.
\end{assum}
Under Assumption~\ref{assum:compensated_model} and also supposing that the robot experiences no contact forces, \eqref{eq:robot_dynamics} results in %
\begin{equation}\label{eq:compensated_dynamics}
    \vM(\vq) \ddot{\vq} = \tilde{\vtau}_J, 
\end{equation}
where $\tilde{\vtau}_J$ is the motor torque for joint control, given by
\begin{equation}
    \tilde{\vtau}_J:=\vtau_J-\vn(\vq, \dot{\vq}),
\end{equation}
with 
\begin{equation}
    \vn(\vq, \dot{\vq}) = \vC(\vq, \dot{\vq}) \dot{\vq} + \vtau_{f} + \vg(\vq).
\end{equation}

The most important physical constraints to consider for real robot arms are the minimum and maximum angular movement, motor velocity, acceleration, torque, and torque derivative limits at each joint
\begin{gather}\label{eq:constraints}
\begin{array}{ll}
\vq_{\min} \leq \vq \leq \vq_{\max}, 
& \dot{\vq}_{\min} \leq \dot{\vq} \leq \dot{\vq}_{\max},\\
\ddot{\vq}_{\min} \leq \ddot{\vq} \leq \ddot{\vq}_{\max},
& \dddot{\vq}_{\min} \leq \dddot{\vq} \leq \dddot{\vq}_{\max},\\
\tilde{\vtau}_{\min} \leq \tilde{\vtau}_{J} \leq \tilde{\vtau}_{\max},
& \dot{\tilde{\vtau}}_{\min} \leq \dot{\tilde{\vtau}}_{J} \leq \dot{\tilde{\vtau}}_{\max},
\end{array}
\end{gather}

Typically, $\dot{\vq}_{\min}\mathord{=}-\dot{\vq}_{\max},\, \ddot{\vq}_{\min}\mathord{=}-\ddot{\vq}_{\max},\,\text{and}\,\dddot{\vq}_{\min}\mathord{=}-\dddot{\vq}_{\max}$. Note that however, the torque limits (hence, also the torque derivative limits) are not necessarily symmetric, \ie $\vtau_{\min} \ne -\vtau_{\max}$. Even when they are equal, compensating for $\vn(\vq, \dot{\vq})$ in \eqref{eq:compensated_dynamics} can asymmetrically reduce the available motor torque for joint control. To simplify the analysis, in the following manuscript discussions we assume conservative symmetric compensated torque and torque derivative bounds.\looseness=-1

\subsection{Problem statement}
In this work we seek a controller that minimizes the braking time  $T_b := t_f - t_0$, where $t_0$ is the braking instant and $t_f$ is the instant when the robot finishes the braking. 
In other words, we aim to find the braking controller $\vu^*$ that solves the following time-optimal control problem:
\begin{equation}\label{eq:ocp}
    \vu^*(t) = \mathrm{arg} \min _{\vu(t)} \int_{t_0}^{t_f} 1 \, dt
\end{equation}
subject to the dynamics in \eqref{eq:compensated_dynamics}, constraints in \eqref{eq:constraints} and the boundary conditions
\begin{align}
    \begin{array}{ll}
    \vq(t_0) = {\vq}_0, & \vq(t_{f}) \ \mathrm{is\, free}, \\ 
    \dot{\vq}(t_0) = \dot{\vq}_0, & \dot{\vq}(t_{f}) = \vnull, \\
    \ddot{\vq}(t_0) = \ddot{\vq}_0, & \ddot{\vq}(t_{f}) = \vnull, \\
    \tilde{\vtau}_J(t_0) = \tilde{\vtau}_{J,0}, & \tilde{\vtau}_J(t_{f}) = \vnull\,,
    \end{array}
\label{eq:initial_conditions}
\end{align}  
where ${\vq}_0$, $\dot{\vq}_0$, $\ddot{\vq}_0$ and $\tilde{\vtau}_{J,0}$ are, respectively, the joint position, velocity, acceleration and torque control input of the robot at the time of braking.

Due to the highly nonlinear inertial coupling through $\vM(\vq)$, the aforementioned optimal control problem can not be solved analytically in general, while solving it numerically is, in general, infeasible for real-time control.  
In the next section we propose a solution that, though is suboptimal, can be employed for real-time control and it further enables the prediction of the robot's behavior during braking. 

\section{The proposed braking scheme}
\label{sec:proposed_scheme}
\begin{figure}[t]
	\centering
	\includegraphics[width=1\columnwidth]{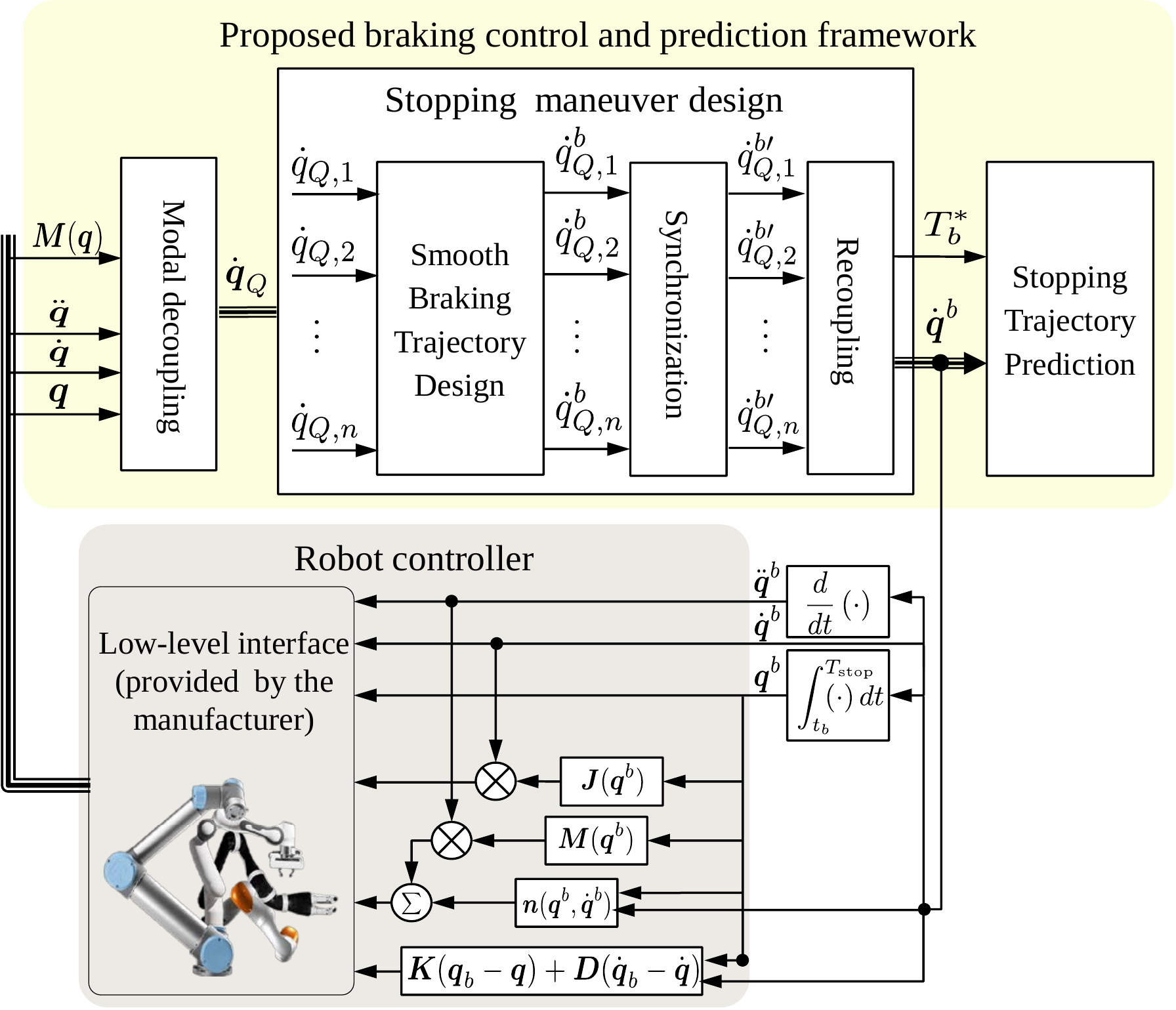}
	\caption{Braking control and prediction architecture.}
	\label{fig:braking_control_architecture}
\end{figure}

Since the motion of each robot joint is highly coupled, our approach obtains a linear transformation that decouples the compensated robot dynamics \eqref{eq:compensated_dynamics} at the acceleration level as soon as the braking signal is raised. This decoupling is later used to transform the initial joint velocities of the system into the modal space. The braking trajectories are geometrically designed using feasible conservative actuator limits to approach zero velocity identically. Moreover, the smoothness of each decoupled velocity braking profile is parameterized such that different braking trajectories for each modal coordinate can be achieved. This feature addresses the difference in joint motion rates while braking and allows a synchronized stopping, favoring a more natural braking behavior. In other words, it ensures that all the joint velocities reach zero simultaneously. The algorithmic pipeline of the proposed approach is illustrated in Fig.~\ref{fig:braking_control_architecture}.

\subsection{Dynamics modal decoupling}

To simplify the analysis of \eqref{eq:compensated_dynamics} we proceed in two steps. First, we linearize this dynamical equation at the braking state ${\vq}_0$. Second, we exploit the positive-definiteness of the inertia matrix to decouple the linearized dynamics.

More precisely, since $\vM({\vq}_0)$ is (symmetric) positive definite, an orthogonal matrix $\vQ \in \mathbb{R}^{n \times n}$ that decouple the dynamics in \eqref{eq:compensated_dynamics} can be always found such that%
\begin{equation}
\begin{aligned}
    \vM({\vq}_0) = \vQ \vM_Q \vQ^{\tp} \,,
\end{aligned}
\label{eq:decoupling_matrix} 
\end{equation}
where $\vM_Q \in \mathbb{R}^{n}$ is the resulting diagonal mass matrix with positive real eigenvalues of $\vM({\vq}_0)$ as its diagonal elements. 
Plugging \eqref{eq:decoupling_matrix} in \eqref{eq:compensated_dynamics} results in system dynamics with a diagonal form
\begin{equation}\label{eq:decoupled_compensated_dynamics}
    \vM_Q \ddot{\vq}_Q = \tilde{\vtau}_{Q},
\end{equation}
where $\ddot{\vq}_Q$ are the new coordinates in the \textit{modal space}, given by
\begin{equation}\label{eq:coordinate_Q_acc}
    \ddot{\vq}_Q = \vQ^{\tp} \ddot{\vq},
\end{equation}
and $\tilde{\vtau}_{Q}$ are the \textit{modal torques}, defined as 
\begin{equation}\label{eq:modal_torque} 
    \tilde{\vtau}_{Q} = \vQ^{\tp} 
    \tilde{\vtau}_J = [\tilde{\vtau}_{Q,1} \ \tilde{\vtau}_{Q,2} \ \cdots \ \tilde{\vtau}_{Q,n}]^{\tp}.
\end{equation}

However, one disadvantage of such a decoupling approach is the introduced coupling in the input torques $\tilde{\vtau}_{Q}$ and their respective constraints in \eqref{eq:constraints}. By restricting the limits of $\tilde{\vtau}_{Q}$ to be independent from each other, the coupling of these transformed input torques can be avoided. However, finding the maximal decoupled limits for  $\tilde{\vtau}_{Q}$, that when transformed back results in optimal feasible torques $\tilde{\vtau}_{J}$ in the original space, usually requires solving another optimization problem \citep{mansfeld2016maximal}. Alternatively, conservative bounds for the torques of the decoupled space can be found analytically. Assuming one obtained either of those, in the following we describe how the braking can be achieved using the modal transformation \eqref{eq:decoupled_compensated_dynamics}--\eqref{eq:modal_torque}.
\subsection{Modal space braking concept}
The overall modal braking idea is based on the physical concept that an inertial object with mass ${\small m_{Q,i}}$ and moving with velocity ${\small \dot{\vq}_Q}$ generates a momentum ${\small m_{Q,i}\:\dot{\vq}_Q}$ along its motion direction (Fig.~\ref{fig:control_regions}(a)). To stop the robot as fast as possible, all joint velocities ${\small \dot{\vq}}$ must be reduced to zero in the shortest time possible. 
For this, the decoupled velocity and position vectors can be obtained by integrating both sides of \eqref{eq:coordinate_Q_acc} over time. For velocity, this results in 
\begin{equation}\label{eq:decoupling_vel_pos} 
    \dot{\vq}_Q = \vQ^{\tp} \dot{\vq}.
\end{equation}
Since $\vQ^{\tp}$ is invertible, the stopping condition of the robot (that is, $\dot{\vq} = \vnull$) is equivalent to the derivative of the vector of decoupled coordinates $\dot{\vq}_Q$ being zero, which can be interpreted as the braking of the vector of decoupled coordinates. Thus, braking in the original space is equivalent to braking the vector of decoupled coordinates. 

To brake each decoupled coordinate as fast as possible, all of the available torque has to be used to reverse its motion direction and then maximally decelerate until zero velocity. Due to the introduced coupling of the torques in \eqref{eq:modal_torque}, not all actuator limits will be hit. For minimum-time braking in the modal space, an optimal combination of modal torque constraints, that also results in feasible maximal torques in the original space, must be used. Obtaining the decoupled torque limits via optimization will inevitably introduce more computation burden which destroys the real-time capability, sacrificing the braking trajectory predictability. To avoid this, we propose an algorithmic scaling approach that results in decoupled yet conservative, sub-optimal modal torque bounds with guaranteed feasibility in the original space. 
The original and modal control regions defined by the physical actuator limits are described next. Then, a systematic analytical solution method for generating conservative (hence, sub-optimal) but independent input torques is introduced. The generated torques are not only within the feasible original control region, but they also enable simultaneous braking for all the robot joints.
\subsection{Physically-admissible braking via momentum scaling}
\label{subsec:modal_braking}

\begin{figure}[t]
    \centering
    \includegraphics[width=1.0\columnwidth]{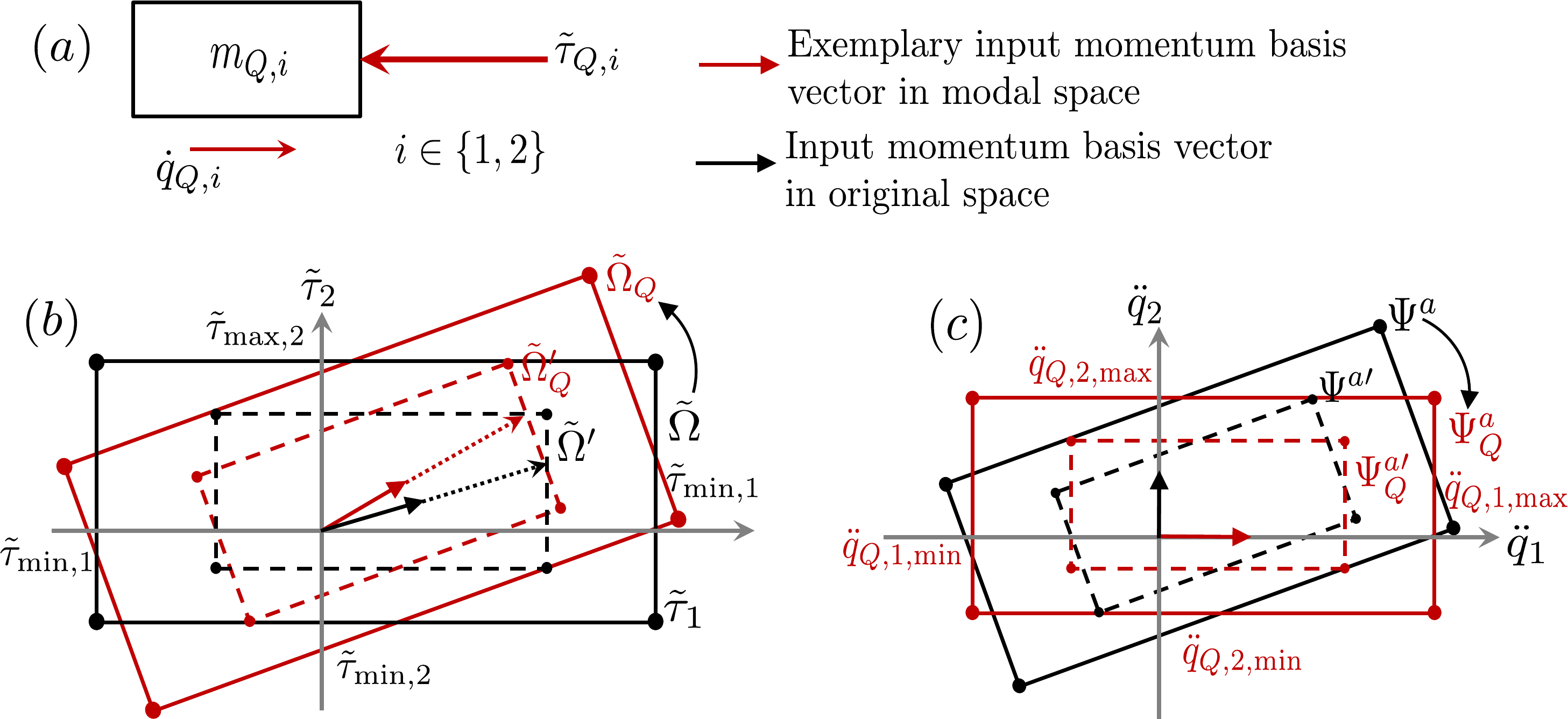}
    \caption{Design of braking control in modal space. (\textit{a}) Intuitive braking torque direction for a mass moving in the modal space (for $n=2$). (\textit{b}) Feasible control torques in the original space ${\small \tilde{\Omega}}$ and modal space ${\small \tilde{\Omega}_{Q}}$. (\textit{c}) Corresponding feasible acceleration regions (${\small {\Psi}^a}$ and ${\small {\Psi}^a_{Q}}$) obtained via scaling. %
    The regions ${\small \tilde{\Omega}^{\prime}/{\Psi}^{a\prime}}$ and ${\small \tilde{\Omega}_{Q}^{\prime}/ {\Psi}_{Q}^{a\prime}}$ denote conservative regions, in which torque/acceleration control inputs stay admissible as specified by the physical actuator limits.}
    \label{fig:control_regions}
\end{figure}

Upon transforming the system dynamics from the original space into the decoupled modal space, the maximum/minimum bounds for the available joint torques (after compensating for gravity, Coriolis and centrifugal torques) are no longer mutually independent \citep{mansfeld2014reaching}. As a result, it is not always feasible, with respect to the actuator limits, to apply any desired torque arbitrarily in the modal space. Since the decoupling property of the modal space allows solving the braking control problem analytically, one has to make sure that the inputs ${\small \tilde{\vtau}_{Q}}$ designed in modal space lie within the admissible available control region. This region can be obtained via e.g.~scaling \citep{mansfeld2016maximal}, which results in conservative modal control inputs within ${\small \tilde{\Omega}_{Q}^{\prime}}$ when represented in the original space (cf.~Fig.~\ref{fig:control_regions}(b)). 

Another way to design braking control inputs utilizing the modal space while ensuring being within the physical actuator limits is introduced next. In the decoupled space, where each state is controlled independently, braking the whole robot is equivalent to stopping ${\small m_{Q,i}}$ as fast as possible. To achieve this, one has to apply a proportional torque ${\small \tilde{\vtau}_{Q,i}}\,;\ i=1,...,n$ in the opposite direction (Fig.~\ref{fig:control_regions}(a)). A safe starting point is to use a counter momentum with equal magnitude, but this has to be scaled up for faster braking. The extent to which we can scale up this braking momentum vector to generate a feasible safe control torque can be determined by searching for the minimum ratio between torque limit and modal mass for all the decoupled coordinates (Fig.~\ref{fig:control_regions}(b)). 
This can be done in real-time using the conservative control bounds scaling algorithm (Algorithm~\ref{alg:conservative_decoupled_modal_limits}) as described in \cite{mansfeld2014reaching}, which we adapted to also evaluate limits for modal-space velocity, acceleration and jerk trajectories (Fig.~\ref{fig:control_regions}(c)). 

As the final braking time depends on the slowest descending state, there is no need to apply higher torques on the other states to stop them sooner. Therefore, we seek simultaneous braking action in which each decoupled state control is synchronized such that all the robot joints are stopped simultaneously. In this case, the time-to-stop that is the same for all coordinates due to simultaneous braking can be estimated from
\begin{align}
T_\text{stop} &= \max_{i=1,\ldots,n} \left( t_{\text{stop},i} :=\frac{m_{Q,i} \dot{\vq}_Q}{\tilde{\vtau}_{Q,i}} \right).
\label{eq:time_to_stop} 
\end{align}
 
The maximum admissible torques available for braking and their corresponding momentum can be obtained via Algorithm~\ref{alg:conservative_decoupled_modal_limits}. The parameters ${\small k_v, k_a, k_j}$ are, respectively, the scaling factors of velocity, acceleration and jerk limits for simultaneous braking, whereas ${\small \left(\vv_{Q,\mathrm{min}}^{'},\vv_{Q,\mathrm{max}}^{'}\right), \left(\va_{Q,\mathrm{min}}^{'},\va_{Q,\mathrm{max}}^{'}\right), \left(\vj_{Q,\mathrm{min}}^{'},\vj_{Q,\mathrm{max}}^{'}\right)}$ are the corresponding modal velocity, acceleration and jerk limits. We denote the approach based on Algorithm~\ref{alg:conservative_decoupled_modal_limits} by \textit{\textbf{consv. scaling}} throughout the manuscript. Note that the proposed controller is of bang-bang type, which may result in oscillatory behavior. %

To further reduce the braking time, optimization techniques can be used to search for maximal \textit{modal bounds}\footnote{These are the limits on the trajectory bounds after applying the decoupling transformation $\vQ$ to the original space.} for designing decoupled velocity braking profiles. For this, we adopt the scheme proposed by \cite{mansfeld2016maximal} that maximizes the volume enclosed by the modal control limits uniformly in all quadrants.\footnote{See \cite{mansfeld2016maximal} for more elaborated details.} The approach for designing modal braking velocity profiles using these optimized limits is denoted by \textit{\textbf{opt. scaling}} in the following. It is noteworthy here that, to use optimized, maximal modal limits while still offering real-time control capability, carrying out optimizations online must be avoided.

\begin{rem} 
  One approach to achieve this is by sampling the reachable workspace of the given robot with desired granularity, such that it is possible to evaluate all its possible joint-space configurations (as was done in \ie \cite{mansfeld2018safety}). Then, the corresponding mass matrices and their modal decompositions can be evaluated symbolically. The required static optimizations to search for the optimal modal limits are carried out offline, which can be stored in a database against the corresponding robot configurations. Based on this data, a \emph{braking map} from the robot configuration to the maximal modal constraints could be learned using strategies similar to the one proposed in \cite{KussabaEtAl2023Learningoptimalcontrollers}. At run time, this map could be queried for the maximal modal constraint.
\end{rem}

Next, we augment our proposed modal braking approach with a parameterized smoothing approach that enables switching from the current velocity profile to the braking profile. This is achieved while respecting the actuator and joint-space limits, while still preserving the predictability of the resulting braking manoeuvers.

\begin{figure}[t]
  \centering
  \vspace{-0.225cm}
  \scalebox{0.95}{
  \begin{minipage}{1.05\linewidth}
  \begin{algorithm}[H]
\small 
\setstretch{0.95}
\caption{Calculating conservative trajectory limits in modal space via constraint hyperrectangle scaling.}
{\small \textbf{Inputs: } $\vv_\mathrm{max}$, $\va_\mathrm{max}$, $\vj_\mathrm{max}$, $\vM(\vq_b)$, $\dot{\vM}(\vq_b,\dot{\vq}_b)$, ${\tilde{\vtau}}_\mathrm{max}$, ${\dot{\tilde{\vtau}}}_\mathrm{max}$}\\
{\small \textbf{Outputs: }
${\left(\vv_{Q,\mathrm{min}}^{'},\vv_{Q,\mathrm{max}}^{'}\right), \left(\va_{Q,\mathrm{min}}^{'},\va_{Q,\mathrm{max}}^{'}\right), \left(\vj_{Q,\mathrm{min}}^{'},\vj_{Q,\mathrm{max}}^{'}\right)}$
} \\
${\quad \small k_v \leftarrow 1,\  k_a \leftarrow 1,\  k_j \leftarrow 1}$ \\
${\quad \small \va_\mathrm{max}\leftarrow \mathrm{min}\left( \va_\mathrm{max},{\vM^{-1}(\vq_b)\,{\tilde{\vtau}}_\mathrm{max}}\right)}$ \\
${\quad \small \vj_\mathrm{max}\leftarrow \mathrm{min}\left( \vj_\mathrm{max},{\vM^{-1}(\vq_b)\left[{\dot{\tilde{\vtau}}}_\mathrm{max}-{\dot{\vM}(\vq_b,\dot{\vq}_b)}\,\va_\mathrm{max}\right]}\right)}$ \\
${\quad} \Psi^{v}=\left[v_{1, \min }, v_{1, \max }\right] \times \cdots \times\left[v_{n, \min }, v_{n, \max }\right]$, with vertices
$\vv_{i}$ \\
${\quad} \Psi^{a}=\left[a_{1, \min }, a_{1, \max }\right] \times \cdots \times\left[a_{n, \min }, a_{n, \max }\right]$, with vertices
$\va_{i}$ \\
${\quad} \Psi^{j}=\left[j_{1, \min }, j_{1, \max }\right] \times \cdots \times\left[j_{n, \min }, j_{n, \max }\right]$, with vertices
$\vj_{i}$ \\
${\quad}${\small \textbf{for} $i \leftarrow 1$ to ${2^n}$ \textbf{do}} \\
${\quad \quad \small \vv_i^Q \leftarrow \vQ \vv_{i},\  \va_i^Q \leftarrow \vQ \va_{i},\  \vj_i^Q \leftarrow \vQ \vj_{i}}$ \\
${\quad \quad \small z_v \leftarrow \mathrm{argmax}(\big|\vv_i^Q| - \big|\vv_\mathrm{max}\big|})$ \\
${\quad \quad \small z_a \leftarrow \mathrm{argmax}(\big|\va_i^Q\big| - \big|\va_\mathrm{max}\big|})$ \\
${\quad \quad \small z_j \leftarrow \mathrm{argmax}(\big|\vj_i^Q\big| - \big|\vj_\mathrm{max}\big|})$ \\
${\quad \quad }${\small \textbf{if}} ${\small \big|\vv_{i,z_v}^Q\big| \mathord{>} \big|\vv_{\mathrm{max},z_v}\big|}$\  %
${\small \left\{k_v \leftarrow \mathrm{min}\left( k_v,\frac{\big|\vv_{\mathrm{max},z_v}\big|}{\big|\vv_{i,z_v}^Q\big|}\right)\right\}}$\  %
{\small \textbf{end if}} \\
${\quad \quad }${\small \textbf{if}} ${\small \big|\va_{i,z_a}^Q\big| \mathord{>} \big|\va_{\mathrm{max},z_a}\big|}$\  %
${\small \left\{k_a \leftarrow \mathrm{min}\left( k_a,\frac{\big|\va_{\mathrm{max},z_a}\big|}{\big|\va_{i,z_a}^Q\big|}\right)\right\}}$\  %
{\small \textbf{end if}} \\
${\quad \quad }${\small \textbf{if}} ${\small \big|\vj_{i,z_j}^Q\big| \mathord{>} \big|\vj_{\mathrm{max},z_j}\big|}$\  %
${\small \left\{k_j \leftarrow \mathrm{min}\left( k_j,\frac{\big|\vj_{\mathrm{max},z_j}\big|}{\big|\vj_{i,z_j}^Q\big|}\right)\right\}}$\  %
{\small \textbf{end if}} \\
${\quad}${\small \textbf{end for}} \\
${\quad \small \vv_{Q,\mathrm{max}}^{'} \leftarrow k_v \vv_\mathrm{max},\  \va_{Q,\mathrm{max}}^{'} \leftarrow k_a \va_\mathrm{max},\  \vj_{Q,\mathrm{max}}^{'} \leftarrow k_j \vj_\mathrm{max}}$ \\
${\quad \small \vv_{Q,\mathrm{min}}^{'} \leftarrow -\vv_{Q,\mathrm{max}}^{'},\  \va_{Q,\mathrm{min}}^{'} \leftarrow -\va_{Q,\mathrm{max}}^{'},\  \vj_{Q,\mathrm{min}}^{'} \leftarrow -\vj_{Q,\mathrm{max}}^{'}}$
\label{alg:conservative_decoupled_modal_limits}
    \end{algorithm}
  \end{minipage}}
\end{figure}

\subsection{Braking concept in the modal space}

Given the feasible maximal modal bounds, a braking trajectory can be designed at the velocity level. To achieve a minimal braking time, an intuitive approach would be to use the maximum limits of the robot joint actuators. Unfortunately, such an approach is infeasible as it results in discontinuity in acceleration which might result in exceeding the actuator limits. To design smooth curves for the direction reversal motions required for braking, we adapted a method that uses a quintic Bezier \citep{sencer2015curvature}. The Bezier curves are mathematically based on Bernstein polynomials and are frequently used for blending linear motion segments in Computerized Numerical Control (CNC) machines ~\citep{ren2019corner}. While linear motion segments in CNC machinery are always defined before the machining process starts, the cobot's path and its corresponding motion segments are generally not known in advance.\footnote{The braking action of these robots can be triggered at any arbitrary time instant since it depends on the dynamic movement of the obstacles in the robot's workspace or fault triggering.} Thus, the blending of velocity segments with different slopes has to be computed in an online fashion.

\subsection{Designing modal space braking curves}
\begin{figure}
\centering
     \includegraphics[scale=0.38]{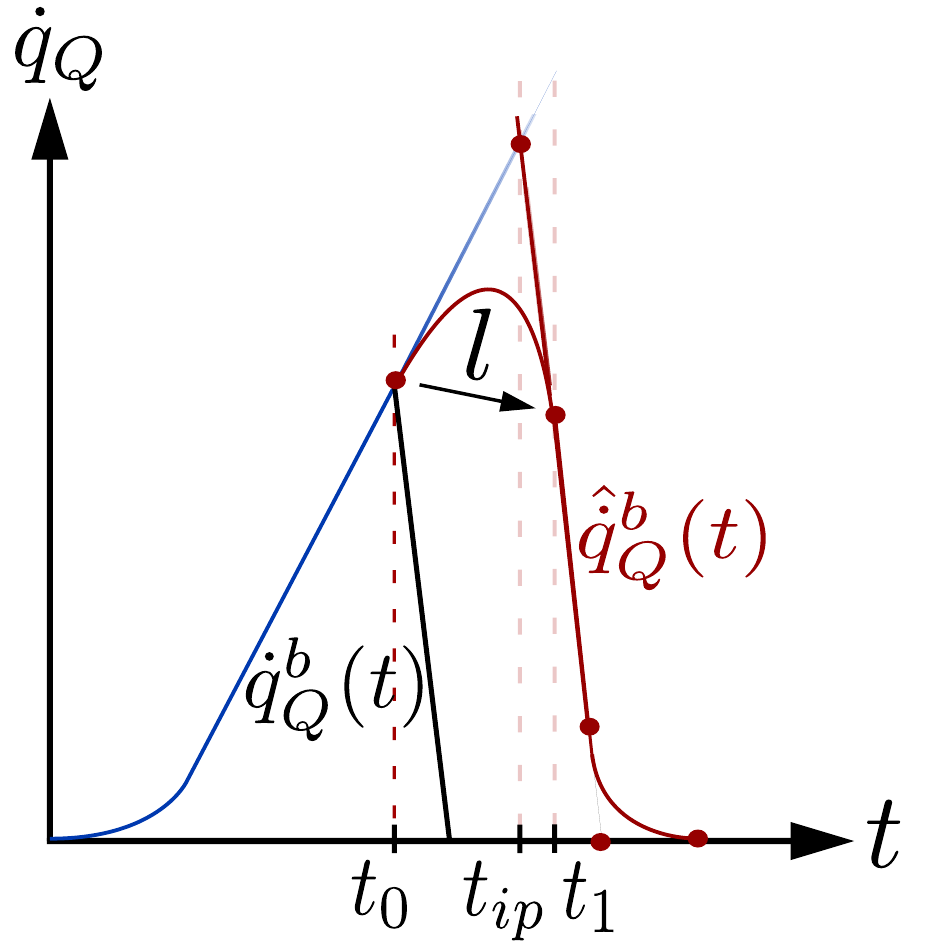} %
     \includegraphics[scale=0.425]{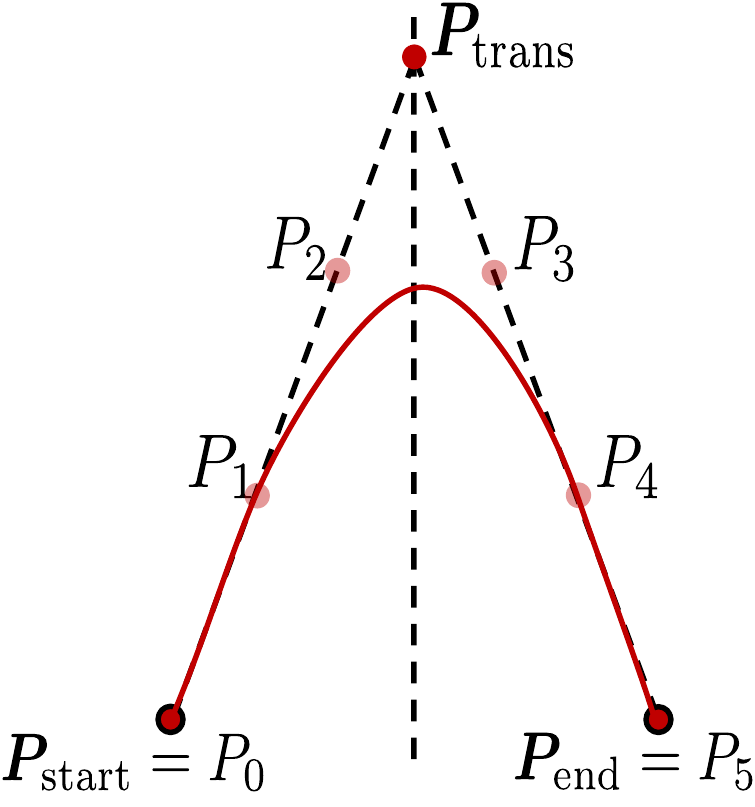} 
\caption{A designed braking curve (\textit{left}) using a quintic Bezier curve (\textit{right}), whose anchor points are given by $P_0 \cdots P_5$ and the corner point is denoted $P_\mathrm{trans}$.}
\label{fig:Brake_Bezier}
\end{figure}

The methodology for designing a smooth decoupled velocity braking trajectory using Bezier curves is summarized graphically in Fig.~\ref{fig:Brake_Bezier}. When the braking velocity signal is raised at time instant $t_0$, each of the components $\dot{q}_{Q,i}(t_0);\ i=1,...,n$ is located already at an inflection point $t_0$. Shifting $t_0$ by $l$ samples results in a new time instant $t_{ip}=t_0+l$ from which the blending to the braking deceleration can be started (Fig.~\ref{fig:Brake_Bezier}, left). The quintic Bezier polynomials use a set of six control points to manipulate the shape of the curve. While these control points can be chosen arbitrarily, in our case they are chosen such that the Bezier curve is symmetric with a $G^2$-smooth transition\footnote{The junction points between $\dot{q}_{Q,i}(t)$ or $\dot{q}^{b}_{Q,i}(t)$ share a common tangent direction and a common center of curvature.} between $\dot{q}_{Q,i}(t)$ and $\dot{q}^{b}_{Q,i}(t)$ (Fig.~\ref{fig:Brake_Bezier}, right). 

Some control points are constrained to satisfy the $G^2$ transition smoothness property. Hence, the Bezier curve for smooth blending can be fully defined by just three points. The three design points are the anchor/corner points defined in $\mathbb{R}^2$ as 
\begin{equation}
    \vP_\mathrm{start} = \begin{bsmallmatrix} t_0 \\ \dot{q}_{Q,i}(t_0) \end{bsmallmatrix},\,
    \vP_\mathrm{trans} = \begin{bsmallmatrix} t_{ip} \\ \dot{q}_{Q,i}(t_{ip})\end{bsmallmatrix},\,
    \vP_\mathrm{end} = \begin{bsmallmatrix} t_{1} \\ \dot{q}^{b}_{Q,i}(t_{1})\end{bsmallmatrix}.
    \label{eq:Beyier_control_points}
\end{equation}

While $\vP_\mathrm{start} $ is known at the braking instant, $\vP_\mathrm{trans}$ and $\vP_\mathrm{end}$ depend on the quantities $\dot{q}_{Q,i}(t_{ip})$ and $\dot{q}^{b}_{Q,i}(t_{1})$ to be estimated (together with the time instants $t_{ip}$ and $t_{1}$). Assuming a constant acceleration at $t_0$, the minimum transition time to a braking velocity $\dot{q}^{b}_{Q,i}(t_e)$ that does not violate the maximum jerk constraint $\dddot{q}_{Q,i,\mathrm{max}}$ can be approximated as detailed next. 

Linearizing $\ddot{q}_{Q,i}(t)$ around $t_0$ using a first-order Taylor series expansion and $\dddot{q}_{Q,i,max}$ gives
\begin{equation}
    \ddot{q}_{Q,i}(t_e)=\ddot{q}^{b}_{Q,i}(t_e) = \ddot{q}_{Q,i}(t_0) + \dddot{q}_{Q,i,\mathrm{max}}(t_e-t_0). \label{eqn:3.15}
\end{equation}

Substituting $\ddot{q}_{Q,i,\mathrm{max}} = \ddot{q}^{b}_{Q,i}(t_e) $, it follows that
\begin{equation}
     \Delta t_{s1} = t_e - t_0 = \frac{\ddot{q}_{Q,i,\mathrm{max}} - \ddot{q}_{Q,i}(t_0)}{\dddot{q}_{Q,i,\mathrm{max}}}. \label{eqn:3.16}
\end{equation} 
 
Linearizing $\dot{q}_{Q,i}(t)$ around $t_0$ using a second-order Taylor series expansion and $\dddot{q}_{Q,i,\mathrm{max}}$ gives
\begin{align}
    \dot{q}_{Q,i}(t_e) &= \dot{q}^{b}_{Q,i}(t_e) \notag \\
    &= \dot{q}_{Q,i}(t_0) + \ddot{q}_{Q,i}(t_0){\Delta t_{s1}} + \frac{1}{2}\dddot{q}_{Q,i,\mathrm{max}}{\Delta t_{s1}^2}.
    \label{eqn:3.17}
\end{align}

Plugging \eqref{eqn:3.16} in \eqref{eqn:3.17} yields $\dot{q}^{b}_{Q,i}(t_e)$, which serves as a starting point for designing $\vP_\mathrm{end}$. Since \eqref{eqn:3.15}--\eqref{eqn:3.17} were computed with a constant acceleration $\ddot{q}_{Q,i}$ assumption, the graph of $\dot{q}_{Q,i}$ is hence simply a straight line with $\ddot{q}_{Q,i}(t_0)$ slope. This means the value of $\dot{q}^{b}_{Q,i}(t_e)$ has to be shifted in time by $h$ samples ($\Delta h$ time units) until the lengths between $\vP_\mathrm{start}$ and $\vP_\mathrm{end}$ with the intersection point of its slopes $\vP_\mathrm{trans}$ are equal. The time instant for blending towards zero velocity is $t_1 = t_e+\Delta h$. Using some trigonometric identities, this problem can be solved analytically, and the blending Bezier curve is thus obtained. The shift in time will relax the $\dddot{q}_{Q,i,\mathrm{max}}$ constraint by increasing the transition time between the accelerations, therefore, reducing the resulting jerk.

Please note that, for each modal velocity curve there will be a total of two inflection points that have to be smoothed. The second one is located at $t_f$ where $\dot{q}^{b}_{Q,i}(t)$ meets the time axis line (i.e.,~the zero velocity line). The other blending curve at $t_f$ is computed analogously. Furthermore, the time spent in the two $G^2$-smooth blends $t_{s,i} = \Delta t_{s1,i} + \Delta t_{s2,i} $ is fully known. 

\subsection{Braking trajectory prediction}
Once the modal braking curves are computed, the whole curve can be re-transformed into the original space. Since it is assumed that the inertia matrix and the decoupling transformation $\vQ$ do not significantly change during braking \citep{mansfeld2014reaching}, using the instantaneous $\vQ$ should result in a close approximation of the real braking velocity trajectory. The overall braking time $T_b$, following the synchronized stopping and smoothing, can be used to compute the curve of joint positions and the end-effector's path in Cartesian coordinates required to estimate the braking trajectory at any time from the braking triggering until the complete stop. Furthermore, the total distance to be covered during braking can also be predicted in advance. 

Using the braking trajectory, the robot's end-effector state $\hat{\vx}^b(t) \in \mathbb{R}^{3+3}$ during braking (encoding its translational position and velocity) can be predicted from
\begin{equation}\label{eq:Cartesian_position_velocity_EE}
\hat{\vx}^b(t) = 
\left[
\begin{aligned}
&\vt({\vq}^b(t)) \\ \,%
&\vJ({\vq}^b(t)) {\dot{\vq}^b}(t)   
\end{aligned}
\right],\ t > t_0,
\end{equation}
where ${}_{\ \ \vnull}^{EE}{\vT}({\vq}^b(t)) = \left(\underset{\small 0\, 0\, 0}{{\vR}({\vq}^b(t))} \ {\Bigg|} \ \underset{\small 1}{{\vt}({\vq}^b(t))}\right) \in \mathbb{R}^{4 \times 4}$ denotes the homogeneous transformation matrix from the robot base frame $\vnull$ to $EE$ with $\vR({\vq}^b(t))$ and $\vt({\vq}^b(t))$ being, respectively, the configuration-dependent rotation matrix and translation vector, while $\vJ_v({\vq}^b(t)) \in \mathbb{R}^{3 \times n}$ is the translational part of the robot's Jacobian matrix. 

\section{Simulation results and discussion}
\label{sec:results}
To test the proposed concepts, the developed braking manoeuvers are applied within our braking control and prediction framework to stop a moving robot in simulation. The considered robot model is a 7-DoF Franka Emika Panda arm comprising an open kinematic chain of rigid bodies connected with revolute joints. The physical actuator and joint-space limits are available online \citep{franka2022robot}. For simplicity and better visibility of the results, we locked all the joints except joints 1, 2 and 4 in all the simulations. 
Furthermore, to compare the proposed braking algorithms with the optimal time-minimizing solution, the optimal braking control problem \eqref{eq:ocp} is formulated as a non-linear program using CasADi \citep{andersson2019casadi} then solved utilizing the Knitro solver \citep{byrd2006knitro}.\looseness=-1 

\begin{figure}[t]
\centering
     \includegraphics[scale=0.9]{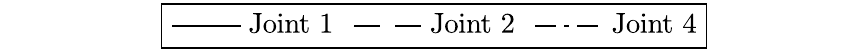} \\
     \includegraphics[scale=1]{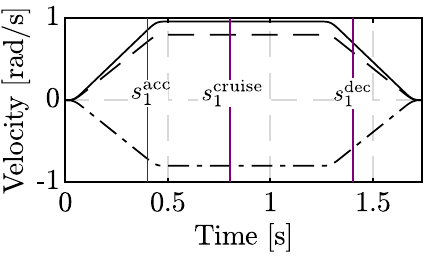}
     \includegraphics[scale=1]{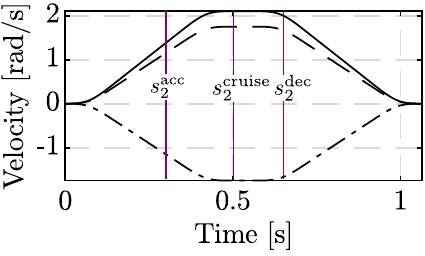}
\caption{Braking settings for robot joints moving with low (\textit{left}) and high (\textit{right}) velocities during different trapezoidal motions. The braking triggering instants $s_i^{\mathrm{X}}$, $i=1,2$ are indicated with purple vertical lines, where the superscript $\mathrm{X} \in \{\mathrm{acc},\mathrm{cruise},\mathrm{dec}\}$ indicates the braking scenario, i.e. ($\mathrm{acc}$) during acceleration, ($\mathrm{cruise}$) during cruise, and ($\mathrm{dec}$) during deceleration. 
}
\label{fig:braking_settings}
\end{figure}

We compared two implementations for designing the smooth braking trajectories of our proposal: 1) Using conservative modal limits (\textit{consv. scaling}), and 2) Using uniform, maximally-optimized feasible ones (\textit{opt. scaling}), following the modal control symmetric limits optimization by volume maximization \citep{mansfeld2016maximal}. Both set of limits were obtained using the decoupling transformation at the braking instant only.\footnote{In addition to the nominal limits from the robot's manufacturer, of course.} Additionally, the braking action of different robot joints was synchronized such that their stopping time matched that of the slowest one. This way the robot does not deviate much from the reference task path, which in turn may not be efficient regarding the kinetic energy dissipation.\footnote{Hence, it maybe less safe upon contact with nearby humans while still in braking.} However, asynchronous braking will result in the same overall braking time. The performance of the stopping actions using our proposed braking manoeuvers designed with the two modal limits scaling approaches is compared against the optimal braking solution. The controllable robot joints are commanded to follow two trapezoidal velocity profiles, each with desired acceleration, cruising, and deceleration phases, as depicted in Fig.~\ref{fig:braking_settings}.
The first reference velocity profile resembles a robot motion with an intermediate velocity, while the maximum joint velocity limits of $\mathord{\sim}2$ m/s are exercised in the second. 

\begin{figure}[t]
\centering
    \includegraphics[scale=0.8]{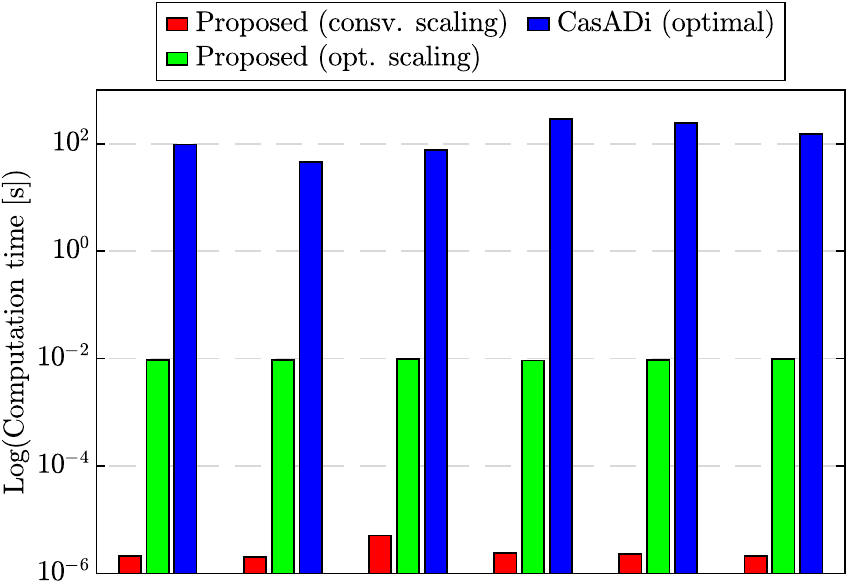}  
     \includegraphics[scale=0.8]{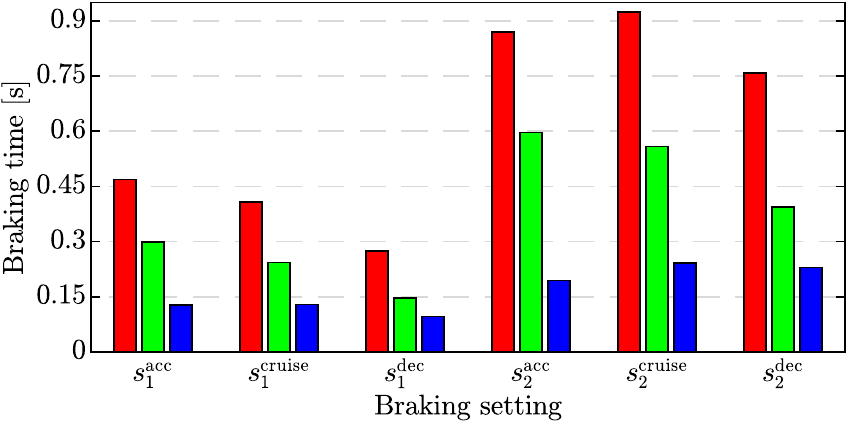} 
\caption{Braking and computation times for braking solutions under different settings.}
\label{fig:computation_braking_times_comparison}
\end{figure}

The braking times and their computation times resulting from applying braking torques generated from the two proposed braking manoeuvers are compared against the optimal solution as depicted in Fig.~\ref{fig:computation_braking_times_comparison}. It can be observed that our proposed manoeuvers require more braking time when the braking action is triggered at the acceleration phase of the trapezoidal motion. Besides the sub-optimality of the modal input limits obtained via both scaling approaches, this is also due to the time invested in smoothly reversing the motion direction. Compared to the acceleration phase, the time it takes to brake during the cruising and deceleration phases is shorter. However, the total time it takes to brake still depends on the speed of the joint at the moment of braking. %

\begin{figure}[t]
\centering
     \includegraphics[scale=0.8]{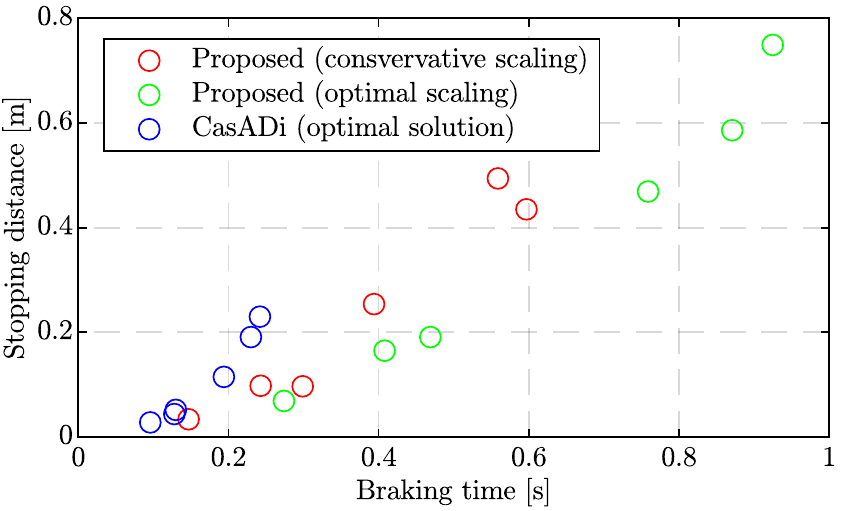}
\caption{Braking times and corresponding stopping distances for braking solutions under different settings.}
\label{fig:braking_distances_times}
\end{figure}

\begin{figure*}[t]
\centering
    {\hspace{0.5cm}\includegraphics[scale=0.98]{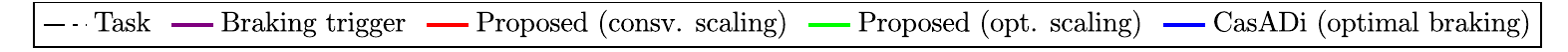}} \\[-0.05cm]
	\begin{tabular}{ c c c }
         {\includegraphics[scale=0.9]{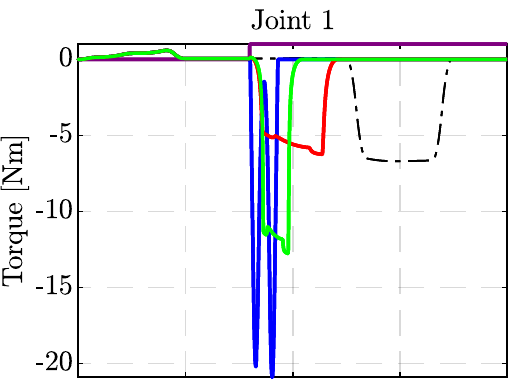}} & {\includegraphics[scale=0.9]{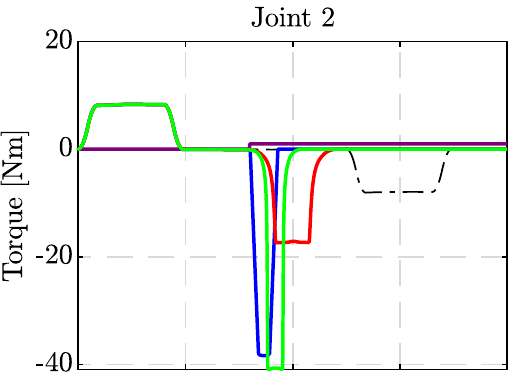}} &
         {\includegraphics[scale=0.9]{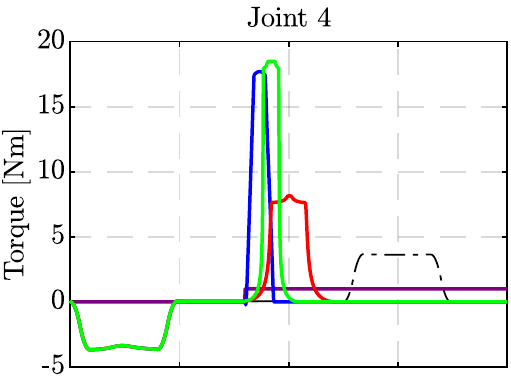}}\\
         {\includegraphics[scale=0.9]{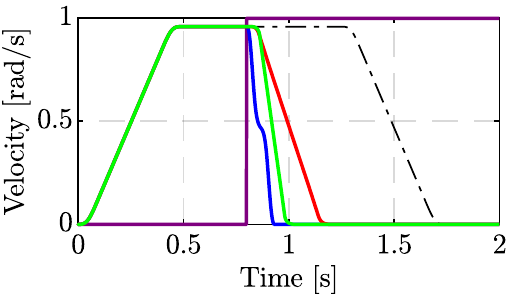}} & {\includegraphics[scale=0.9]{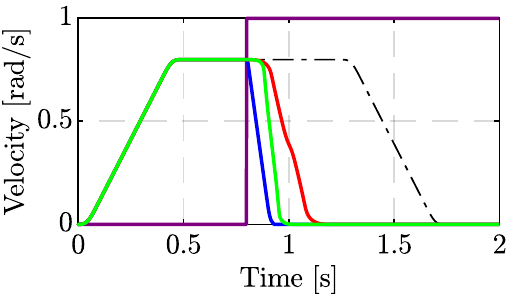}} &
         {\includegraphics[scale=0.9]{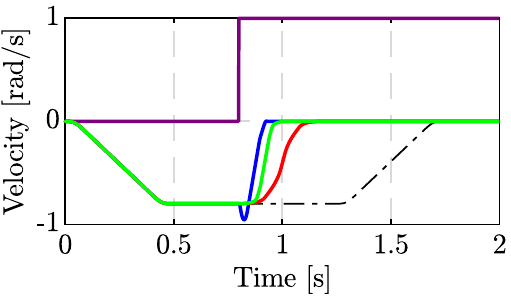}}
    \end{tabular}
\caption{ \centering Robot joint braking control torques (upper row) and resulting joint velocities (lower row) versus time for the sample setting $s^\mathrm{cruise}_1$ (cf.~Fig.~\ref{fig:braking_settings} for the reference trapezoidal motion profile of the task).}
\label{fig:3dof_results_vel_torq}
\end{figure*}

A sample of the applied braking torque and the resulting joint velocity profiles for the braking setting $s^\mathrm{cruise}_1$ is shown in Fig.~\ref{fig:3dof_results_vel_torq}. It can be observed that the desired motor torques always comply with the actuator constraints, while the velocities of all controlled joints converge to zero smoothly without violating their limits. As expected for the employed bang-bang-like control law for braking, one hyperplane of the modal-space control region (input limit) is reached. In the shown sample of Fig.~\ref{fig:3dof_results_vel_torq}, only joint 2 used half of its maximum possible torque.

The braking distance was also evaluated for the two proposed braking manoeuvers against the optimal solution for both braking settings, together with the required braking time, as shown in Fig.~\ref{fig:braking_distances_times}. Recall that this distance depends on the forward kinematics/homogeneous transformation matrix of the robot, which is essentially a nonlinear mapping from its joint configurations to its end-effector's pose in 3D Cartesian space. Obviously, these results confirm that the smaller the braking time, the shorter the stopping distance, consequently decreasing the risk of collisions with the robot. Furthermore, the obtained predicted values for braking distances can be used online for planning robot reactions such as, e.g., collision avoidance or activating safe velocity scaling using the Safe Motion Unit \citep{haddadin2012making}. A comparative summary of key features of the investigated braking methods is provided in Tab.~\ref{tab:comparison of braking controllers}.  

\section{Conclusions and future work}
\label{sec:conclusion}
In this paper, we addressed the problem of braking controlled robot joints while ensuring online and accurate stopping trajectory prediction. We proposed a braking control architecture that relies on designing smooth braking velocity profiles independently in the modal space while taking the actuator and joint-space constraints into account. Two approaches for designing braking manoeuvers in the modal space were developed and integrated into a unified braking control and stopping trajectory prediction framework. The proposed methods were verified on a realistic 7-DoF robotic system with only three moving joints in simulation. The results showed that using the proposed framework yields smooth and fast braking manoeuvers that ensure a predictable stopping behavior of the robot joints. %

For future work, we seek to reduce the braking time  further by using less conservative feasible modal limits. This could be done by maximizing the modal control region volume in each quadrant separately as described in \cite{mansfeld2016maximal}, then the obtained optimal modal bounds can be employed for designing the decoupled braking velocity profiles optimally. %

\begin{table}[t]
\caption{Comparison of braking features.}
\resizebox{\columnwidth}{!}{%
\begin{tabular}{@{}lccc@{}}
\toprule
       \multirow{2}{*}{Braking features}            & CasADi      & Proposed               & Proposed                    \\
                  & (optimal) & (consv. scaling) & (opt. scaling) \\ \midrule
Minimum-time attainment      & \multicolumn{1}{c}{100\%} & \multicolumn{1}{c}{26--31\%}   & \multicolumn{1}{c}{43--51\%}        \\
Stopping distance?  & \multicolumn{1}{c}{\checkmark } & \multicolumn{1}{c}{\checkmark}   & \multicolumn{1}{c}{\checkmark}        \\
Computation time  & \multicolumn{1}{c}{25--32 s} & \multicolumn{1}{c}{$\ll$1 ms}   & \multicolumn{1}{c}{9--10 ms}        \\
Real-time control & $\times$  (offline)             & \checkmark  (real-time)             & \checkmark  (online)                \\ 
Trajectory prediction        & $\times$               & \checkmark            & \checkmark                 \\ \bottomrule
\end{tabular}}%
\label{tab:comparison of braking controllers}
\end{table}

\end{document}